# Analysis and Understanding of Various Models for Efficient Representation and Accurate Recognition of Human Faces

Dharini S, Guru Prasad M, Hari haran. V, Kiran Tej J L, Kunal Ghosh

Department of Computer Science & Engineering , Sir M Visvesvaraya Institute of Technology

Bangalore-562157, Karnataka, India.

Email: face_rec_mvit@googlegroups.com

*Abstract*—**In this paper we have tried to compare the various face recognition models against their classical problems.**
**We look at the methods followed by these approaches and evaluate to what extent they are able to solve the problems.**
**All methods proposed have some drawbacks under certain conditions.To overcome these drawbacks we propose a multi-model approach.**

*Keywords-Face Recognition; Computer Vision; Human Computer Interaction.*

## I. INTRODUCTION

Over the past 20 years numerous face recognition papers have been published in the computer vision community; a survey can be found in [1]. The number of realworld applications (e.g., surveillance, secure access, human/computer interface) and the availability of cheap and powerful hardware also lead to the development of commercial face recognition systems. Despite the success of some of these systems in constrained scenarios, the general task of face recognition still poses a number of challenges with respect to changes in illumination, facial expression, and pose.

## II. EIGEN FACES BASED FACE RCOGNITION

It is one of the holistic approaches of face recognition, which considers the whole face. For a given face database a set of Eigen faces is obtained. Each face in the database can be represented as a linear combination of few of the Eigen faces (not all Eigen faces - dimensionality reduction). The weights corresponding to each of the faces can be compared to recognize faces, instead of the comparing the faces as a whole.

Eigen face approach is based on principal component analysis (PCA)[3]. The basic outline algorithm of Eigen faces can be represented as shown in the figure.

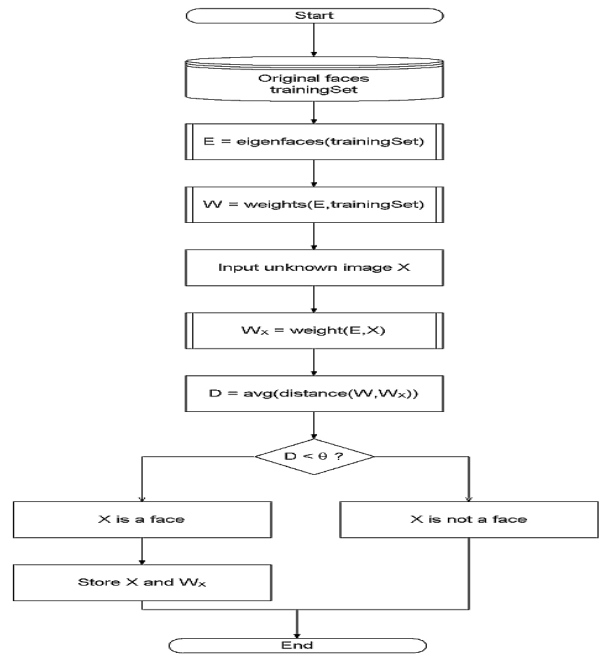

Figure 1:Functioning of Eigen face-based face recognition

The procedure for computing Eigen faces can be mathematically represented as shown below[4]:
- Given face data base of *M* images, Let each face be of size $h\,X\,w$
- Each image is transformed into a vector of size (*hw*) and placed into a set
  $$\{\Gamma_1, \Gamma_2, \cdots, \Gamma_M\}$$
- The adjusted data set is computed by subtracting each vector from the mean

Mean is

$$\Phi_i = \Gamma_i - \Psi$$

- Adjusted data set is obtained as shown below

$$\Psi = \frac{1}{M}\sum_{i=1}^{M}\Gamma_i$$

- The covariance matrix $C \in \mathbb{R}^{D\times D}$ is defined as:

$$C = \frac{1}{M}\sum_{i=1}^{M}\Phi_i\Phi_i^T = AA^T,$$

where,

$$A = \{\Phi_1, \Phi_2, \cdots, \Phi_M\} \in \mathbb{R}^{D\times M}$$

- It is difficult to compute a covariance matrix of *hw X hw,* but this problem can be solved by multiplying $A^TA$ instead of $A^TA$
- The eigenvalue and eigenvector matrices of *C* are

$$U = AV\Lambda^{-1/2}$$

where

$$U = \{u_i\}$$

is the collection of Eigen faces.

The approach for recognition of a new face is described below:[5]
1. Compute a set of weights based on the input image and the Eigen faces by representing the input image as a linear combination of the Eigen faces.
2. To determine if the image is a face at all (whether know or unknown) check if the image is sufficiently close to "face space".
3. If it is a face, classify the weight pattern as either a known person or as unknown.
4. (optional) Update the Eigen faces and/or weight patterns.
5. (optional) If the same unknown face is seen several items, calculate its characteristic weight pattern and incorporate into the known faces

Changing lighting conditions cause relatively few errors, while performance drops dramatically with size change. This is obvious since under lighting changes alone the neighborhood pixel correlation remains high, but under size changes the correlation from one image to another is quite low. It is clear that there is a need for a multi scale approach, so that faces at a particular size are compares with one another.

### III HIDDEN MARKOV MODEL BASED RECOGNITION.

Hidden Markov Models (HMM) have been successfully used for speech recognition and more recently in action recognition where data is essentially one dimensional over time. In this section we investigate the recognition performance of a one dimensional HMM for gray scale face images. For frontal face images the significant facial regions (hair,eyes,nose etc) come in a natural order from top to bottom, even if the images undergo small rotations in the image plane and/or rotations in the plane perpendicular to the image plane. Each of these facial regions is assigned to a state in a left to right 1D continuous HMM.

The state structure of the face model and the non-zero transition probabilities $a_{ij}$ are shown in Figure 2.

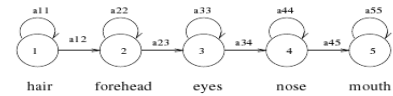

Figure 2: Left to right HMM for face recognition

### III.A HMM FEATURE EXTRACTION.

Each face image of width W and height H is divided into overlapping blocks of height L and width W. The amount of overlap between consecutive blocks is P (Figure 3).

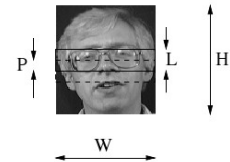

Figure 3: Face image parameterization and blocks extraction

The variations of the recognition performance with parameters P and L is extensively discussed in [2]. However, the performance of the system is less sensitive to variations in L, as long as P remains large $(P \leq L-1)$ .In [2] the observation vectors consist of all the pixel values from each of the blocks, and therefore the dimension of the observation vector is $L \times W$ .The use of the pixel values as observation vectors has two important disadvantages: First, pixel values do not represent robust features, being very sensitive to image noise as well as image rotation , shift or changes in illumination. Second, the large dimension of the observation vector leads to high computational complexity of the training and recognition systems. This can be critical for a face detection or recognition system that operates on a large database or when the recognition system is used for real time applications.

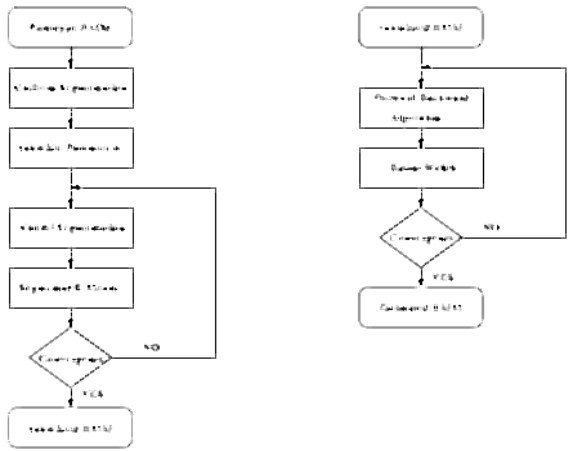

Figure 5: Training Scheme

### III.b Training the face Models

For face detection, a set of face images is used in the training of one HMM. The images in the training set represent frontal face of different people taken under different illumination conditions.

For face recognition, each individual in the database is represented by an HMM face model. A set of images representing different instances of the same face are used to train each HMM.

After extracting the blocks from each image in the training set, the observation vectors (Karhunen-Loeve Transform coefficients) are obtained and used to train each of the HMMs. First, the HMM
$\lambda = (A,B,\pi)$ is initialized. The initial values of A and Pi are set given the left to right structure of the face model. At the next iteration, the uniform segmentation is replaced by the Viterbi segmentation. The iteration stop, and the HMM is initialized, when the Viterbi segmentation likelihood at consecutive iterations is smaller that a threshold. The final parameters of the HMM are obtained using the Baum-Welch recursive procedure.

After testing the algorithm with Olivetti Database an accuracy of 86 % is achieved. We also notice that the HMM based algorithm is more suitable for Frontal faces and occluded face images brings down the accuracy of the algorithm, it is able to gracefully work with lighting and moderate pose variations.

### IV.A Dynamic Link Architecture

DLA is one of the feature based approach implemented for face recognition. Its basically a neutral information processing paradigm.
Although neural networks has the ability to derive meaning from complicated or precise data can be used to extract patterns & detect trends that are too complex for various computer techniques. It requires a *separate perturbation* for facial recognition, so solving of problems related to change in expression of faces is inefficient.

Advantage of this technique is that various application are accessible via DLA such as distortion invariant object recognition, sensory segmentation and scene analysis.
One of the prominent feature of DLA is the use of *synaptic plasticity* i.e ability of connection between two neurons to change in strength. This enables it to instantly group sets of neurons into higher symbolic units. Conventional neural systems do not provide this ability to bind separate subsets of neurons, inevitably merging them into one structure less global assembly.

Recently, a comparative study was performed for three well known face recognition techniques, namely, the Eigen faces, the fisher faces and classification neural networks, and the elastic graph matching [6]. It was found that the Eigen faces worked well when the face images had relatively small lighting and moderate expression variations. Their performance deteriorated significantly, as lighting variation increased. On the contrary, the elastic graph matching was found relatively insensitive to variations in lighting, face position and expressions.

### IV.B Image Representation

In this method we basically deal with two domains image domain I, model domain M. Biologically speaking, I may correspond to primary visual cortical areas, and *M* to infero temporal cortex [7].

Image Domain:
It contains two dimensional array of nodes represented as
$A_x^I = \{(x,\alpha) \mid \alpha = 1,2,---,F\}$
Where x is the position of the node and $\{1,2---,F\}$ are different feature detector neurons [8].
Each node $A_x^I$ contains set of activity signals which are known as Jets (local description of grey value distribution) basically set of feature vectors which are based on Gabor type wavelets [9].
$J_x^I = \{S_{x\alpha}^I \mid \alpha = 1,2---,F\}$

In order to establish dynamic links between nodes in a image domain we determine connections i.e for neuron$(x,\alpha)$ and $(y,\beta)$ represented as $T_{x\alpha;y\beta}^I$
Model Domain:

It's a set of attributed graphs. All being idealized copies of sub graphs in the image domain.

$T^I_{x\alpha;y\beta}$ – represents connection between image and model domain, these connection are feature preserving

If there is a connection between corresponding neurons in corresponding nodes then we say there is a matching graph i.e. *isomorphism*

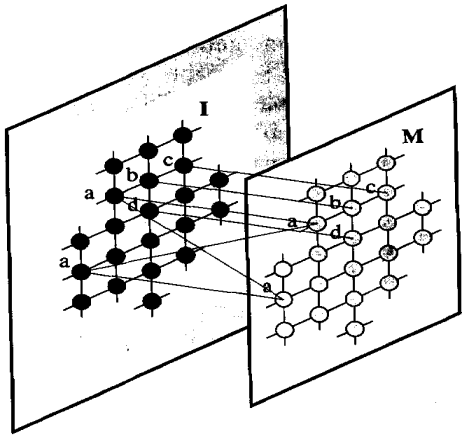

Diagram:

Matching graphs in image domain and object domain. Within the object domain there is a sub graph *M* that is identical to a sub graph *I* in the image domain: *I* and *M* contain the same features *(a, b, c, . . .)* in the same arrangement. In both domains, neurons in neighboring nodes are connected. Connections between domains are feature-type preserving, but not position specific

Basically we use elastic graph matching which has 2 phases
a) Identifying appropriate sub graph I of the full image domain.
b) Identification of the matched sub graph M in the model domain.

### IV.C Various Binding

1. Binding all nodes that belong to same class or object.
2. Neighborhood relationships within the image of the object i.e. between neurons of neighboring nodes. ($T^I_{x\alpha;y\beta}$ where x ≠ x')
3. Binding feature cells within one node onto a jet . ( i.e $T^I_{x\alpha;x\beta}$ within a node)
4. $T^{IM}$ i.e binding between points in I and corresponding points in M.

So we say two graph are identical if there is neighborhood preserving and feature type preserving mapping between almost all nodes of I and M.

### IV.D Implementation

1. A set of feature vectors for a image is determined where feature vectors are based on Gabor type wavelets [9].
2. We represent a image as a graph with vertices corresponding to jets [10] [11].
3. We select a sparse graph and find best match with model graphs using Elastic graph matching algorithm [12].

The Dynamic Link Architecture derives its power from a data format based on syntactically linked structures. This capability has been exploited here on three levels. Firstly, when an image is formed in the image domain, the local feature detectors centered at one of its points are bundled to form a composite feature detector (called a jet). A composite feature detector can be shipped to the model domain and can be compared as a whole to other composite feature detectors there. This frees the system from the necessity to train new individual neurons as detectors for complex features before new object classes can be recognized, a major burden on conventional layered systems. Secondly, links are used to represent neighborhood relationships within the image domain and within the model domain. Neural objects thereby acquire internal structure, and their communication can now be constrained to combinations with matched syntactical structure. This forms the basis for elastic graph matching. Finally, the dynamic binding between matched graphs, which in the present context is an unimportant by-product of recognition, will be useful to back-label the image with all the patterns recognized and to build up representations of composite objects and scenes.

### V. Component Based Method

This can be achieved in many ways. One of the most popular technique is a 3D morph-able model[15]. This was developed by Volker Blanz and Thomas Vetter, who extended the 2D approach of face recognition. The earlier system namely the global face recognition system was based on the whole face pattern. The drawback of such a system was the need of a large number of training images taken from different views and under different lighting conditions. Hence there was a need for a new approach to pose and illumination invariant face recognition. This was accomplished by the component based method of face recognition. 3D morph-able face models apply the general concept into the vector space representation of the face models. The main idea behind this model is that given a sufficiently large database of 3D face models, any arbitrary face can be generated by morphing between the ones in the database.[16] Based on a pair of images of a person's face, the morph-able model allows the computation of a 3D face. When the 3D face models of all the subjects in the training database are computed, arbitrary synthetic face images under varying pose and illumination are generated to train the component based recognition system. The method used to create a 3D model from a set of 2D images is called "Analysis by Synthesis Loop"[7]. This helps to find the parameters such that the rendered images of the 3D model are as close as possible to the input images.

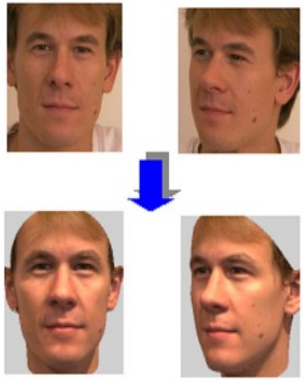

Figure 1: Generation of the 3D model. The top images are the real images used to generate a 3D model. The bottom images are synthetic images generated from the model. We can notice the similarity between the original and synthetic images[16]

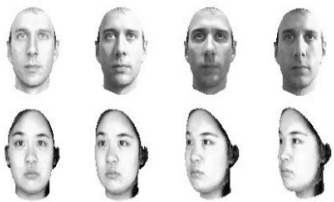

Figure 2: Synthetic training images. Synthetic face images generated from the 3D head models under different illuminations (top row) and different poses (bottom row) [16]

Feature vectors are generated during the face detection phase. The test face will be trained on these feature vectors in a one Vs all approach. The normalized outputs of various component classifiers will be compared. The highest value of the output of a identity will be taken to be the identity of the test face.

ADVANTAGES

1. The flexible positioning of the components can compensate for changes in the pose of the face.
2. Those parts of the face which do not contain relevant information can be omitted by choosing only the appropriate components.

RESULTS

Results on 1200 real images of six subjects show that the component-based recognition system clearly outperforms a comparable global face recognition system.

The resulting ROC curves of global and component-based recognition on the test set can be seen in Figure 3. The component-based system achieved a recognition of 90%, which is approximately 50% above the recognition rate of the global system[16].

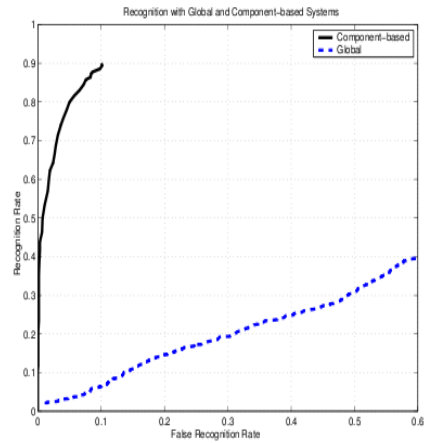

Figure 3: ROC curves for the component-based and the global face recognition systems.
Both systems were trained and tested on the same data.

## VI. FISHER FACES

Moses etal., "the variations between the images of the same face due to illumination and viewing direction are almost always larger than image variations due to change in the face identity".
This is where the need for another classical method ,Fisher faces which is an example of class specific method comes to picture. This means it tries to shape the scatter in order to make it more reliable for classification. This method selects W in such a way that a ratio of the between class scatter and within class scatter is maximize.[13]

$$r = \frac{S_b}{S_w}$$

i.e. is max.

$$S_B = \sum_{i=1}^{c} N_i (\mu_i - \mu)(\mu_i - \mu)^T$$

$$S_W = \sum_{i=1}^{c} \sum_{x_k \in X_i} (x_k - \mu_i)(x_k - \mu_i)^T$$

] $W_{opt}$ is chosen as the matrix with orthonormal columns which maximizes the ratio of the determinant of the between class scatter matrix of the projected samples to the determinant of the within class scatter matrix of the projected samples.

$$W_{opt} = \arg \max_{W} \frac{|W^T S_B W|}{|W^T S_W W|}$$

$$= [w_1 \ w_2 \ \ldots \ w_m]$$

where $\{w_i | i = 1, 2, \ldots, m\}$

is the set of generalized eigenvectors of $S_b$ and $S_w$ corresponding to the m largest generalized eigenvalues. $\{\lambda_i | i = 1, 2, \ldots, m\}$, i.e.,

$$S_B w_i = \lambda_i S_W w_i, \quad i = 1, 2, \ldots, m$$

There can be utmost c-1 Eigen (generalized) values on m where c is the number of classes.

what is the problem of this method?

$S_W \in \mathbb{R}^{n \times n}$ is always singular.

This stems from the fact that the rank of $S_w$ is utmost n-c and in general the number of images in the learning set N is much smaller than the number of pixels in each image n. This means that it is possible to choose the matrix N such that the within class scatter of the projected samples can be made exactly zero.

To overcome the complication of the singular $S_w$ an alternate criteria is proposed called fisher faces avoids this problem by projecting the image set to a lower dimensional space. So that the resulting within class scatter matrix $S_w$ is non singular. This is achieved by the PCA to reduce the dimension of the feature space to N-C and then applying the standard FLD defined by (4) to reduce the dimensions to c-1.

More formally $W_{opt}$ is given by,

$$W_{pca} = \arg \max_{W} |W^T S_T W|$$

$$W_{fld} = \arg \max_{W} \frac{|W^T W_{pca}^T S_B W_{pca} W|}{|W^T W_{pca}^T S_W W_{pca} W|}$$

The second method chooses W to maximize the between class scatter. For ex. The second method which we are currently investigating chooses W to maximize the between class scatter of the projected class after first reduce the between class scatter. Taken to an extreme, we can maximize the between-class scatter of the projected samples subject to the constraint that the within class scatter is zero, i.e.

where $W_{opt} = \arg \max_{W \in \mathcal{W}} |W^T S_B W|$

is the set of n X m matrices with orthonormal columns contained in the kernel of $S_w$.

The problems like the illumination is overcome in this method and the result of the experiment conducted by the researches on the YALE database is given above which shows that the error rate has reduced by the considerable amount.[13]

TABLE 1
COMPARATIVE RECOGNITION ERROR RATES FOR GLASSES/
NO GLASSES RECOGNITION USING THE YALE DATABASE

| Glasses Recognition | | |
|---|---|---|
| Method | Reduced Space | Error Rate (%) |
| PCA | 10 | 52.6 |
| Fisherface | 1 | 5.3 |

VI. CONCLUSIONS

The above discussed methods do not solve all the challenges faced by face recognition individually. So there is a need for a novel method that can overcome all the problems. In the multi-model method that we propose, first we identify the properties of the test image to be recognized. The properties include variation of the face pose from frontal face( by finding the L2 norm of the test face from a representative frontal face (using eigen faces)), changing intensity of the light incident on the face, degree of face occlusion, facial expression etc. Depending on the degree to which the test image has these properties we select an appropriate face recognition model.